\documentclass[conference]{IEEEtran}
\IEEEoverridecommandlockouts
\usepackage{cite}
\usepackage{amsmath,amssymb,amsfonts}
\usepackage{graphicx}
\usepackage{textcomp}
\usepackage{xcolor}
\usepackage{bm}
\usepackage{amsmath}
\usepackage{amssymb}
\usepackage{amsthm}
\usepackage{mathrsfs}
\usepackage{enumerate}
\usepackage{multirow}
\usepackage{color}
\usepackage{threeparttable}
\usepackage{subfigure}

\usepackage[square, comma, sort&compress, numbers]{natbib}
\usepackage[pagebackref=false,breaklinks=true,letterpaper=true,colorlinks,bookmarks=false]{hyperref}
\usepackage{times}
\usepackage{breakurl}
\usepackage{array}
\usepackage{verbatim}
\usepackage{algorithm}
\usepackage{bbding}
\usepackage{algpseudocode}
\usepackage{lettrine}

\usepackage{booktabs}
\usepackage{cleveref}
\usepackage{pifont}
\usepackage{pgfplots}
\usepackage{fontawesome}

\def\BibTeX{{\rm B\kern-.05em{\sc i\kern-.025em b}\kern-.08em
    T\kern-.1667em\lower.7ex\hbox{E}\kern-.125emX}}
\begin{document}

\title{Efficient Feature Fusion for UAV Object Detection\\ 

}

\author{\IEEEauthorblockN{Xudong Wang}
\IEEEauthorblockA{\textit{School of Computer Science
} \\
\textit{East China Normal University}\\
Shanghai, China \\
71265901006@stu.ecnu.edu.cn}
\and
\IEEEauthorblockN{Yaxin Peng}
\IEEEauthorblockA{\textit{Department of Mathematics \&} \\
\textit{School of Future Technology}\\
\textit{Shanghai University}\\
Shanghai, China \\
yaxin.peng@shu.edu.cn}
\and
\IEEEauthorblockN{Chaomin Shen\textsuperscript{\faEnvelope}
}
\IEEEauthorblockA{\textit{School of Computer Science} \\
\textit{East China Normal University}\\
Shanghai, China \\
cmshen@cs.ecnu.edu.cn}
}

\maketitle

\begin{abstract}
Object detection in unmanned aerial vehicle (UAV) remote sensing images poses significant challenges due to unstable image quality, small object sizes, complex backgrounds, and environmental occlusions. Small objects, in particular, occupy small portions of images, making their accurate detection highly difficult. Existing multi-scale feature fusion methods address these challenges to some extent by aggregating features across different resolutions. However, they often fail to effectively balance the classification and localization performance for small objects, primarily due to insufficient feature representation and imbalanced network information flow. In this paper, we propose a novel feature fusion framework specifically designed for UAV object detection tasks to enhance both localization accuracy and classification performance. The proposed framework integrates hybrid upsampling and downsampling modules, enabling feature maps from different network depths to be flexibly adjusted to arbitrary resolutions. This design facilitates cross-layer connections and multi-scale feature fusion, ensuring improved representation of small objects. Our approach leverages hybrid downsampling to enhance fine-grained feature representation, improving spatial localization of small targets, even under complex conditions. Simultaneously, the upsampling module aggregates global contextual information, optimizing feature consistency across scales and enhancing classification robustness in cluttered scenes. Experimental results on two public UAV datasets demonstrate the effectiveness of the proposed framework. Integrated into the YOLO-v10 model, our method achieves a 2 percentage points improvement in average precision (AP) compared to the baseline YOLO-v10 model, while maintaining the same number of parameters. These results highlight the potential of our framework for accurate and efficient UAV object detection.
\end{abstract}


\section{Introduction}
In recent years, unmanned aerial vehicles (UAVs) have gained substantial attention due to their versatility and wide-ranging applications~\cite{ref58}. Among these, UAV-based object detection has emerged as a critical task with significant implications in fields such as agriculture, disaster management, military surveillance, and urban planning. This task involves automatically identifying and localizing objects of interest—such as vehicles, buildings, or vegetation—in images captured by UAV-mounted cameras, enabling efficient decision-making in complex scenarios.

UAV-based object detection presents unique challenges compared with traditional ground-based object detection. The aerial perspective introduces unstable image quality, small object sizes, complex backgrounds, and frequent environmental occlusions. Small objects, in particular, occupy only a small fraction of the image, further complicating accurate detection. These challenges necessitate innovative solutions to address the limitations of existing object detection frameworks.

\begin{figure}[t]
    \centering
    \includegraphics[width=1\linewidth]{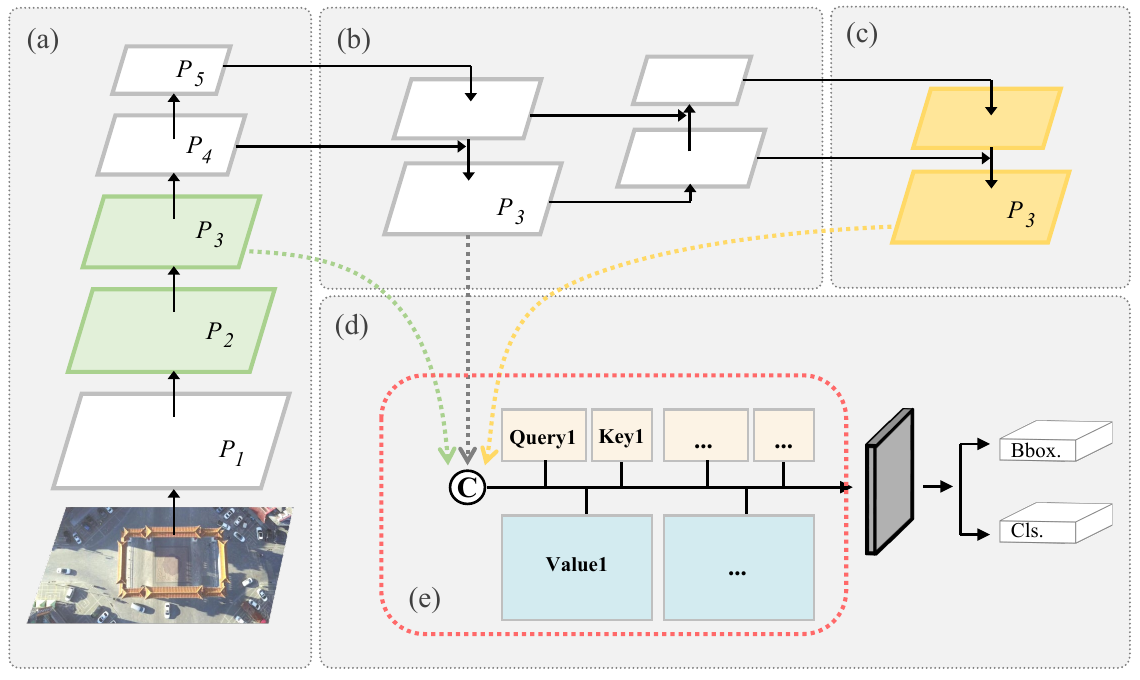}
\caption{Illustration of our framework. (a) Backbone with fusion down sample (FDS) module added for collecting shallow layer features. (b) Neck of network. (c) Fusion up sample (FUS) module for collecting deeper layer features. (d) Detection head. (e) Fusion multi-head self-attention (FMSA) module for the feature fusion of outputs from (a), (b), and (c).}
\label{fig:Introduction}
\end{figure}

Currently, deep learning techniques are the dominant approach for UAV-based object detection~\cite{ref61,ref62,ref63,ref64,ref65}. Existing methods can be categorized into three main groups. 
The first group includes two-stage detectors, such as Faster R-CNN, which use a region proposal network (RPN) to generate candidate regions, followed by classification and regression. 
These models are known for their high accuracy but often suffer from slower inference speeds, limiting their applicability in real-time scenarios. 
The second group comprises single-stage detectors, such as the YOLO series, which directly predict object categories and bounding boxes in a single step. 
While these models offer faster inference, they face challenges in detecting small objects due to insufficient feature representation. The third group consists of Transformer-based detectors, such as Vision Transformer (ViT) and Detection Transformer (DETR), which excel at modeling global features and large receptive fields. However, their computational complexity increases significantly when processing high-resolution UAV images, and their performance on small object detection often falls short compared to CNN-based methods. Object detection~\cite{ref68,ref71,ref72} has recently seen advancements driven by diffusion models~\cite{ref46, ref39,ref73, ref74}, which have demonstrated remarkable capabilities in generating high-resolution synthetic data and enhancing model robustness in challenging scenarios. These models are widely applied in domains such as autonomous driving, medical imaging, and remote sensing, where precise detection and localization are critical.

One prominent challenge for object detection algorithms, e.g., mainstream CNN-based ones, is the imbalance between the classification and localization information. As network architectures become deeper, the enhancement of semantic features often leads to a degradation of spatial details. Additionally, detection heads in multi-scale architectures, such as YOLO-v10, are limited in their ability to balance high-resolution and low-resolution features, resulting in redundant structures and reduced localization accuracy, particularly for small objects.

To address these challenges, we propose a novel framework for feature fusion that integrates cross-layer connections and multi-scale fusion. As illustrated in Fig.\,\ref{fig:Introduction}, our method seamlessly integrates into existing CNN-based architectures. By leveraging Fusion Down Sampling (FDS) and Fusion Up Sampling (FUS) modules, it captures valuable features from shallow and deep network layers. These features are fused with the network’s output using a Fusion Multi-Head Self-Attention (FMSA) module, significantly enhancing the detection performance in UAV scenarios. Specifically, the FDS module enhances fine-grained feature representation, which is crucial for improving the spatial localization of small targets, even under complex conditions. Meanwhile, the FUS module aggregates global contextual information, optimizing feature consistency across scales and enhancing classification robustness in cluttered scenes. This dual approach improves both localization accuracy and classification performance. Our framework is designed to be flexible and can be easily integrated into various CNN architectures, making it a comprehensive enhancement for UAV object detection tasks.

The main contributions of this work are summarized as follows:
\begin{itemize}
    \item[\textcolor{black}{$\bullet$}] We design a feature fusion framework with full-scale and full receptive fields to address the imbalance between classification and localization information in UAV-based object detection.
    \item[\textcolor{black}{$\bullet$}] We propose a hybrid downsampling module, which enhances the network's ability to extract and utilize shallow-layer features. This enables efficient cross-layer feature fusion during the downsampling process and improves information flow to deeper layers.
    \item[\textcolor{black}{$\bullet$}] We introduce a hybrid upsampling module, leveraging global attention to improve object-background separation and the detection of occluded objects, enhancing overall detection robustness.
    \item[\textcolor{black}{$\bullet$}] The proposed framework is compatible with existing CNN-based models. By supporting hybrid upsampling and downsampling, it enables feature maps at various network depths to be flexibly adjusted to arbitrary resolutions, facilitating long-range cross-layer connections and multi-scale feature fusion.
\end{itemize}

\section{Related Work}\label{sec:rw}

\subsection{Traditional Object Detection Methods}
Traditional object detection algorithms rely on handcrafted feature extraction and sliding window techniques. These methods typically involve three key stages: region proposal, feature extraction, and classification regression. During the region proposal stage, potential object locations are identified. Then, handcrafted methods are  employed to extract features from these candidate regions, followed by classification using traditional classifiers. Representative approaches include the Viola–Jones detector~\cite{ref12} and the Histograms of oriented gradients (HOGs) for human detection~\cite{ref13}. While these methods laid the groundwork for object detection, they suffer from high computational complexity, limited feature representation capabilities, and optimization difficulties, making them unsuitable for modern UAV detection tasks.


\begin{figure*}[t]
    \centering
    \includegraphics[width=1\textwidth]{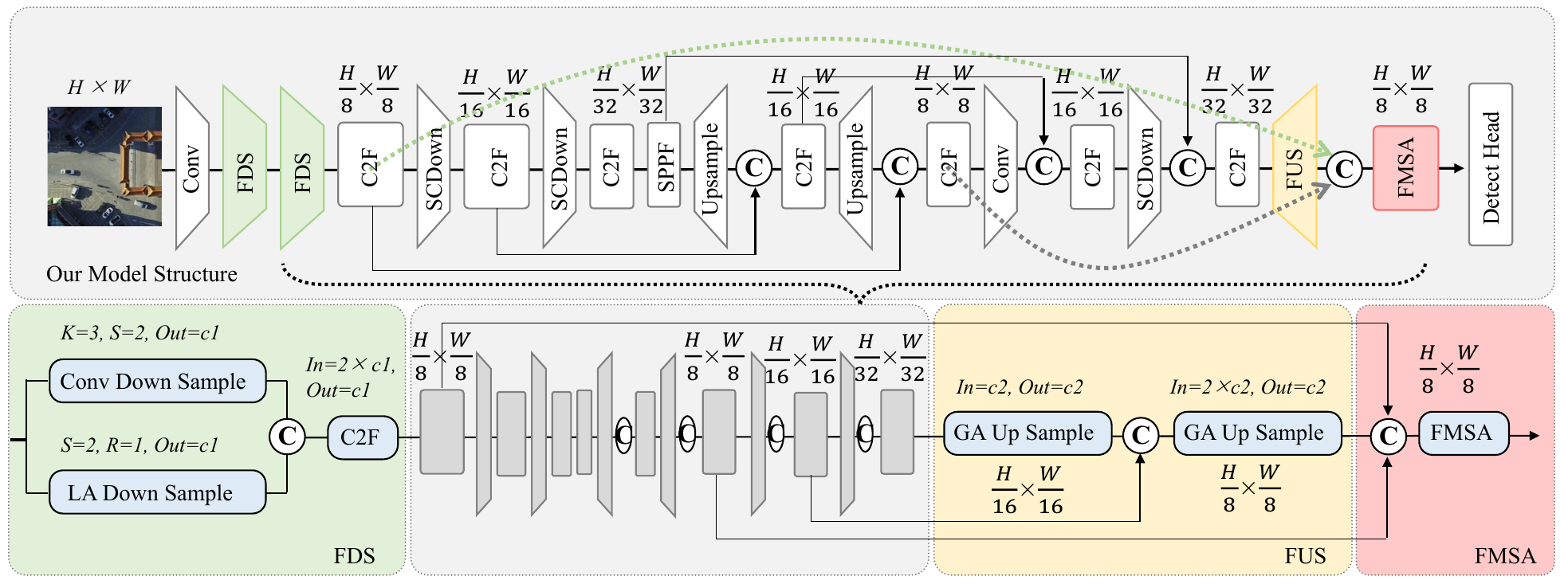}
\caption{Illustration of our model architecture. The proposed method is a supplementary component and integrated to the state-of-the-art model YOLO-v10, including FDS, FUS, FMSA modules.}
\label{fig:Structure}
\end{figure*}

\subsection{Deep Learning-Based Object Detection}
Deep learning has revolutionized object detection by leveraging CNNs. These approaches can be broadly categorized into two-stage and single-stage detection methods. 

Two-stage detectors, such as Faster R-CNN~\cite{ref15}, utilize a region proposal network (RPN) to generate candidate regions, followed by classification and regression. Extensions like RecFRCN~\cite{ref79} and Improved Mask R-CNN~\cite{ref80} improve multi-scale feature representation and segmentation capabilities. Despite their high accuracy, two-stage models often require significant computational resources, making them less suitable for real-time applications.

In contrast, single-stage detectors, such as YOLO-v8~\cite{ref52} and FasterNet-SSD~\cite{ref75}, predict object categories and bounding boxes in a single pass, enabling faster inference. Improved versions of YOLO (e.g., YOLO-v9~\cite{ref56} and YOLO-v10~\cite{ref48}) and other models like DTSSNet~\cite{ref55} and A2Net~\cite{ref77} have further enhanced detection performance. However, these methods struggle with detecting small or occluded objects, leading to higher false positive rates, especially in UAV scenarios.

Recent advances in Transformer-based object detection have introduced powerful architectures for global feature modeling. Vision Transformer (ViT)~\cite{ref35, ref82, ref83} applies the Transformer architecture to computer vision, offering larger receptive fields and flexible weight-sharing strategies. Swin Transformer~\cite{ref36} employs a hierarchical structure with a sliding window approach, while DETR~\cite{ref38} frames object detection as a set prediction task, enabling end-to-end detection. 

\subsection{UAV Object Detection}
While Transformer-based detectors have shown promising global modeling capabilities, their direct application to UAV imagery remains challenging. Specifically, these models face challenges in handling high-resolution UAV images due to excessive computational overhead and suboptimal performance for small objects.
UAV object detection presents additional challenges, such as small object sizes, complex backgrounds, and environmental occlusions. Multi-scale feature learning~\cite{ref18,ref60} has proven effective for addressing these issues. Hybrid approaches combining CNNs and Transformers, such as Conformer~\cite{ref39} and LPSW~\cite{ref42}, leverage local feature extraction with CNNs and global feature modeling with Transformers. Other methods, including DIAG-TR~\cite{ref43} and NeXtFormer~\cite{ref81}, enhance the representation of small objects by integrating convolutional layers with Transformer blocks.

Despite these advancements, most existing methods overlook the imbalance between classification and localization, particularly for UAV-specific detection tasks. The method proposed in this paper addresses this imbalance through an advanced feature fusion framework. By leveraging cross-layer connectivity and multi-scale fusion, our approach effectively enhances detection performance while reducing redundant structures. Furthermore, the proposed method is seamlessly compatible with existing CNN-based architectures, making it highly practical for UAV object detection scenarios.



\section{Proposed Method}\label{sec:method}
This section introduces our proposed fusion multi-head self-attention (FMSA) framework, designed to address the challenges of UAV object detection by enhancing the balance between classification and localization information, particularly for small object detection. The FMSA framework integrates two auxiliary modules, FDS and FUS, to perform multi-scale feature fusion and long-range cross-layer connections. 

\subsection{Overall Architecture}
The proposed FMSA module acts as a supplementary component for CNN-based networks and is seamlessly integrated into the data flow of the baseline YOLO-v10 model. As shown in Fig.\,\ref{fig:Structure}, the architecture incorporates the FDS and FUS modules, which adjust feature maps from different network depths to resolutions of \(H/8 \times W/8\), where they are fused using global multi-head self-attention. 

\begin{figure}[t]
    \centering
    \includegraphics[width=1\linewidth]{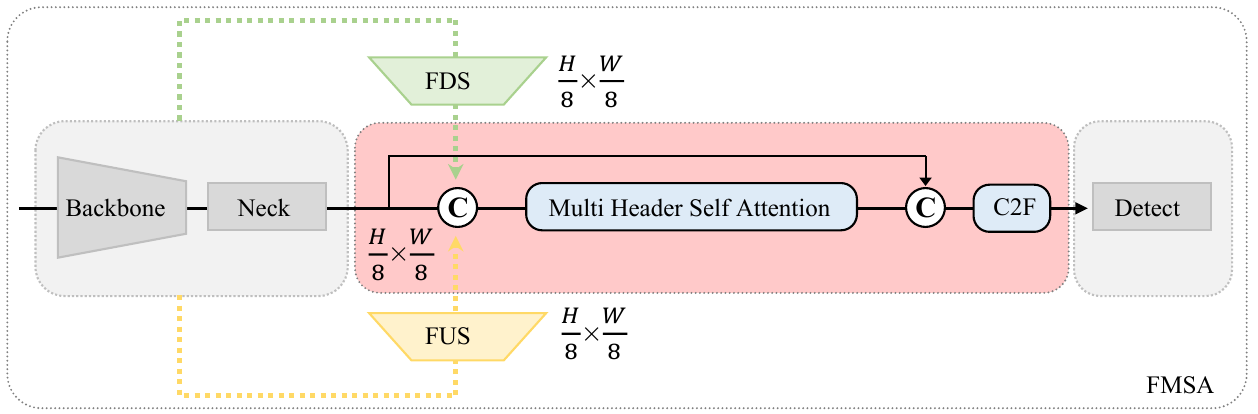}
\caption{Illustration of FMSA module. It is an additional module to CNN-based network. Apart from the main output of the network, the FDS module collects shallow layer features, and the FUS module collects deeper layer features for feature fusion conducted by the FMSA module.}
\label{fig:FMSA}
\end{figure}

Specifically, the FDS module replaces the second and third convolutional downsampling layers in YOLO-v10, capturing valuable shallow-layer features with higher resolutions. Towards the end of the network, the FUS and FMSA modules are introduced just before the detection head. The FMSA module collects outputs from the FDS, FUS, and the original network to enhance feature fusion and refine localization accuracy. The detection heads for large and medium objects are disabled to prioritize small-object detection, a common challenge in UAV scenarios.

\subsection{Fusion Multi-Head Self-Attention (FMSA)}
Object detection tasks require effective processing of both classification and localization information, but deep networks often experience an imbalance between these two aspects. This issue is exacerbated in small-object detection, where semantic information from deeper layers cannot be effectively balanced with spatial information from shallower layers. 

To address this, we propose the FMSA module, which aggregates multi-resolution features from different network depths using global multi-head self-attention (MHSA). By combining high-resolution features from shallow layers and semantically rich low-resolution features from deeper layers, the FMSA module improves the network's capacity for detecting small objects while maintaining localization precision.

As illustrated in Fig.\,\ref{fig:FMSA}, the FMSA module consists of multi-head self-attention (MSA) blocks, a multi-layer perceptron (MLP), and residual connections. Positional information is retained through standard sinusoidal position embeddings. The input feature map \(\bm x\) is reshaped into \(H \times W\) tokens with dimension \(C\), represented as:
\[
\bm{x}_p, \quad p \in \{1, 2, \dots, H \times W\}. \tag{1}
\]
The self-attention block computes the output as:
\[
\bm{Z}_0 = \left[\bm{x}_1; \bm{x}_2; \dots; \bm{x}_p\right] + \bm{E}_{pos}, \quad \bm{E}_{pos} \in \mathbb{R}^{(H \times W) \times C}, \tag{2}
\]
where \(\bm{E}_{pos}\) denotes positional embeddings. Subsequent layers are computed as:
\[
\bm{Z}_L' = \operatorname{MSA}(\operatorname{LN}(\bm{Z}_{L-1})) + \bm{Z}_{L-1}, \tag{3}
\]
\[
\bm{Z}_L = \operatorname{MLP}(\operatorname{LN}(\bm{Z}_L')) + \bm{Z}_L'. \tag{4}
\]
The final output of the FMSA module is given by:
\[
\bm{y} = \operatorname{C2F}(\operatorname{MHSA}(FDS_{out} + FUS_{out} + \bm{x}) + \bm{x}), \tag{5}
\]
where C2F is a YOLO-v10's module for effective feature extraction and fusion.

\subsection{Fusion Down Sample (FDS)}
In CNN-based object detection, shallow layers often lack attention despite containing high-resolution spatial information crucial for small-object detection. The FDS module enhances the downsampling process by introducing the local attention downsample (LADS) operation, which captures spatial details more effectively than traditional convolutional downsampling.

\begin{figure}[t]
    \centering
    \includegraphics[width=1\linewidth]{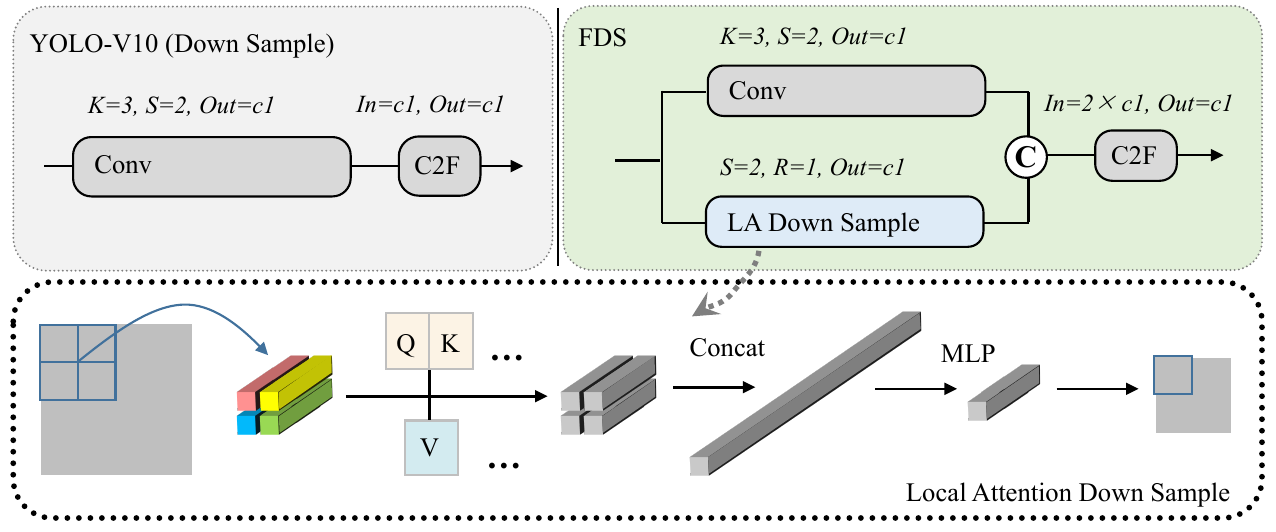}
\caption{Illustration of FDS module. On the right is the network structure of the FDS module, and on the left is the downsampling module of YOLO-v10 for comparison.}
\label{fig:FDS}
\end{figure}

As shown in Fig.\,\ref{fig:FDS}, the LADS module applies self-attention to 2×2 patches, aggregates outputs through an MLP, and reshapes them to a reduced resolution. For input feature map \(\bm x\):
\[
\bm{Z}_{local'} = \operatorname{PatchExtract}(\bm{x}, \text{stride, stride}),\tag{6}
\]
\[
\quad \bm{Z}_{local'} \in \mathbb{R}^{\text{(stride} \times \text{stride)} \times C}.\tag{7}
\]
\[
\bm{Z}_{local} = \operatorname{MLP}(\operatorname{MSA}(\operatorname{LN}(\bm{Z}_{local'}))), \quad \bm{Z}_{local} \in \mathbb{R}^{1 \times C}.\tag{8}
\]
The FDS output is computed as:
\[
\bm{y} = \operatorname{C2F}(\operatorname{Concat}(\operatorname{Conv}(\bm{x}), \bm{Z})).\tag{9}
\]
This module enables shallow-layer features to flow into deeper layers, improving small-object detection by preserving high-resolution spatial details.

\subsection{Fusion Up Sample (FUS)}
In UAV object detection, accurate small-object detection benefits from high-resolution feature maps. The FUS module incorporates global self-attention mechanisms to upsample lower-resolution feature maps while preserving global semantic information. 

\begin{figure}[t]
    \centering
    \includegraphics[width=1\linewidth]{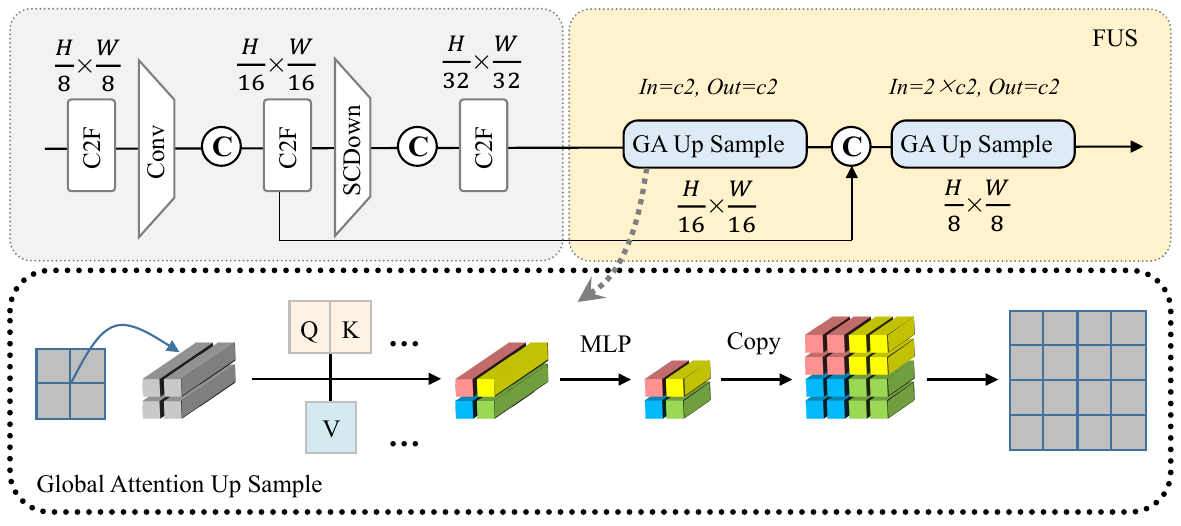}
\caption{Illustration of FUS module. It performs up-sampling operation for the target deeper layers.}
\label{fig:FUS}
\end{figure}

As shown in Fig.\,\ref{fig:FUS}, feature maps from resolutions \(H/32 \times W/32\) and \(H/16 \times W/16\) are processed through global attention upsample (GAUS) blocks to produce \(H/8 \times W/8\) feature maps. The GAUS operation is defined as:
\[
\bm{Z} = \operatorname{Replicate}(\operatorname{MLP}(\operatorname{MSA}(\operatorname{LN}(\bm{x})))),\tag{10}
\]
where $\bm{x} \in \mathbb{R}^{(H \times W) \times C}$,
and $\bm{Z} \in \mathbb{R}^{(H \times W \times \text{stride} \times \text{stride}) \times \frac{C}{\text{stride}}}$.

The transformed feature maps are then fused with the FDS output and the original network features in the FMSA module, improving detection accuracy by enhancing the distinction between objects and background, as well as handling occlusions effectively.

The FUS module contributes to improving feature representation consistency and better localization for small objects, ensuring robust detection in UAV imagery.


\section{Experiment and Analysis}\label{sec:exp}  
In the experiment, two public datasets, the VisDrone2019 dataset and the DOTA-v1.5 dataset, were used to train and test the proposed method.

\subsection{Datasets}
\textbf{\emph{VisDrone2019}:}
This dataset is a large-scale benchmark collection created by the AISKYEYE team at the Lab of Machine Learning and Data Mining, Tianjin University, China. It contains carefully annotated ground truth data for various computer vision tasks related to drone-based images. The dataset contains 10,209 static images from diverse scenes and environments, with 6,471 images designated as the training set and 548 images used for validation. The data was collected using various drone platforms under different scenes, weather, and lighting conditions, and covers common objects such as pedestrians, cars, bicycles, and tricycles.\\
\indent  \textbf{\emph{DOTA-v1.5:}}
DOTA is a large-scale dataset for object detection in aerial imagery. It is designed for the development and evaluation of object detectors in aerial images, featuring objects with various scales, orientations, and shapes. DOTA-v1.5 consists of 2,806 high-resolution aerial images from diverse scenes and environments, with annotations also provided for extremely small instances (smaller than 10 pixels). Of these, 1,411 images are used as the training set and 458 images as the validation set. The dataset includes 15 common object categories, such as large vehicles, small vehicles, ships, bridges, and airplanes.

\subsection{Evaluation Metrics}  
In object detection metrics, most references for small object detection in UAV-based images use mean average precision (mAP) as the evaluation metric~\cite{ref53,ref54,ref47}. Here average precision (AP) is a numerical measure that summarizes the shape of the Precision-Recall curve, used to determine the optimal balance between precision and recall in practice. Therefore, in the comparative experiments, mAP is used as the primary metric, as it provides a comprehensive assessment of the model's performance. The definition of mAP is as follows:
\[m A P=\frac{1}{n} \sum_{i=1}^{n} A P_{i}. \tag{11} \]
Here $n$ represents the number of identified object categories, and $AP_{i}$ denotes the area under the Precision-Recall curve for the $i$-th object category.

\subsection{Implementation Details} 
In this paper, we conduct experiments on the Ubuntu 22.04 system with two NVIDIA GeForce RTX 3090 GPUs, each equipped with 24GB of memory. The experimental environment includes PyTorch 2.0.1, CUDA 11.7, and Python 3.9. We evaluate our method on the VisDrone2019 and DOTA-v1.5 datasets, which contain 8k training images and 1k validation images in total. During training, the input feature map dimensions are set to 640 × 640, and the model is trained for a maximum of 1000 epochs. The batch size is 6, and we use 8 workers for data loading. YOLO-v10 is used as the baseline model, with our modules applied to it. The model is trained using the SGD optimizer with a momentum of 0.937, and the initial learning rate is set to 0.01. We adopt the default training configuration of YOLO-v10 with the specific parameter settings as shown in TABLE \ref{table:Configuration}.
\begin{table}[t]
    \centering
    \caption{Network Parameters and Computer Configurations}
    \begin{tabular}{@{}l|l|l@{}}
        \hline
        \textbf{Category} & \textbf{Parameter} & \textbf{Value} \\ 
        \hline
        Network Parameters & Optimizer & SGD \\
        & Learning rate (lr0) & 0.01 \\
        & Momentum & 0.937 \\
        & Batch size & 6 \\
        & Maximum Epochs & 1000 \\
        & Image size & 640 × 640 \\ 
        \hline
        Computer Configurations & Operating System & Ubuntu 22.04 \\
        & CPU & 3.30GHz \\
        & GPU & GeForce RTX 3090 \\
        & RAM & 64.0 GB \\ 
        \hline
    \end{tabular}
    \label{table:Configuration}
\end{table}
\subsection{Comparison with State-of-the-art Methods} 
\subsubsection{Comparisons on VisDrone2019}
\begin{table*}[t]
\centering
\caption{Detection performance (\%) of different methods on VisDrone2019 dataset.}
\begin{tabular}{c||cc||cc}
\hline \textbf{ Model } & \quad$\mathbf{mAP_{50}}$ & $\mathbf{mAP_{50-90}}$ & \,\textbf{ Parameters (M) } & \textbf{ FLOPs (G) } \\
\hline \text { SSD ~\cite{ref32} } & 23.9 & 13.1 & 24.5 & 87.9 \\
\text { RetinaNet ~\cite{ref33} } & 27.3 & 15.5 & 19.8 & 93.7 \\
\text { ATSS ~\cite{ref49} } & 31.7 & 18.6 & 10.3 & 57.0 \\
\text { Faster-RCNN ~\cite{ref15} } & 33.2 & 17.0 & 41.2 & 207.1 \\
\text { GCGE-YOLO ~\cite{ref51} } & 34.1 & 19.2 & 4.5 & 10.8 \\
\text { YOLOv5s ~\cite{ref50} } & 35.4 & 20.5 & 9.1 & 23.8 \\
\text { \quad\quad Swin Transformer ~\cite{ref36} \quad\quad } & 35.6 & 20.6 & 34.2 & 44.5 \\
\text { YOLOv8s ~\cite{ref52} } & 36.4 & 21.6 & 11.1 & 28.5 \\
\text { C3TB-YOLOv5 ~\cite{ref53} } & 38.3 & 22.0 & 8.0 & 19.7 \\
\text { TPH-YOLO ~\cite{ref54} } & 39.3 & 23.6 & 51.5 & 138.1 \\
\text { DTSSNet ~\cite{ref55} } & 39.9 & 24.2 & 10.1 & 50.4 \\
\text { YOLOv9 ~\cite{ref56} } & 43.4 & 26.5 & 51.0 & 239.0 \\
\text { LV-YOLOv5 ~\cite{ref57} } & 41.7 & 25.6 & 36.6 & 38.8 \\
\text { YOLOv10x ~\cite{ref48} } & 46.2 & 28.8 & 31.7 & 171.1 \\
\hline \textbf{ Ours } & \textbf{48.3} & \textbf{30.4} & 31.2 & 490.9 \\
\hline
\end{tabular}
\label{table:VisDrone_Result}
\end{table*}
To validate the effectiveness of the proposed method in the context of small target detection in drone image scenes, a series of comparative experiments were conducted on the VisDrone2019 dataset. Since the same public datasets were used, we referenced the experimental data from a recently published paper~\cite{ref47}, which include some classic two-stage models, one-stage models, and several recently proposed state-of-the-art models. In addition, we selected the latest state-of-the-art model, YOLO-v10, and conducted comparative experiments under similar numbers of model parameters and training settings.\\
\indent To perform a comprehensive performance comparison, our baseline comparison model is YOLOv10x, the largest model in the YOLO-v10 series. For the sake of comparability and to maintain similar parameter scales, we integrated our module into the YOLOv10m, a medium-sized model. TABLE \ref{table:VisDrone_Result} shows that on the VisDrone2019 dataset, our detection model achieves a $mAP_{50}$ of 48.3\%, which represents a 2.1\% improvement over the YOLOv10x model. Overall, the results demonstrate that our model significantly improves small target detection accuracy.
\subsubsection{Comparisons on DOTA-v1.5}
In addition to the VisDrone2019 dataset, we have specifically selected the DOTA-v1.5 dataset as our evaluation dataset. The DOTA-v1.5 dataset is designed for the development and evaluation of object detectors in aerial images, featuring objects of various scales, orientations, and shapes. It includes annotations for extremely small instances (smaller than 10 pixels). The high-resolution aerial images, coupled with small and dense target instances, increase the difficulty and challenges of detection. By validating on the DOTA-v1.5 dataset alongside the VisDrone2019 dataset, we can effectively demonstrate the performance of the proposed method in detecting small objects.

To further validate the effectiveness of the proposed method on the DOTA-v1.5 dataset, our module was integrated into the YOLOv10n, a smaller model in the YOLO-v10 series. TABLE \ref{table:DOTA_Result} shows that on the DOTA-v1.5 dataset, under the same number of model parameters, our detection model achieves a mAP50 of 42.5\%, which represents a 2.0\% improvement over the YOLOv10n model.
\begin{table}[t]
\centering
\caption{Detection performance (\%) on DOTA-v1.5 dataset.}
\begin{tabular}{c||cc||c}
\hline \textbf{ Model } & $\mathbf{mAP_{50}}$ & $\mathbf{mAP_{50-90}}$ & \textbf{ Parameters (M) } \\
\hline \text { YOLOv10n } & 40.5 & 24.9 & 4.5  \\
\textbf{ Ours } & \textbf{42.5} & \textbf{26.7} & 4.5 \\
\hline
\end{tabular}
\label{table:DOTA_Result}
\end{table}

\subsection{Ablation Studies and Analysis}  

In this section, we use VisDrone2019 as the benchmark dataset and YOLOv10n as the baseline model. To ensure that the total number of parameters remains approximately the same across all experimental methods, we integrate our module into the baseline model while simultaneously reducing the number of its network layers. This allows us to observe the effects of each module and the issue of imbalance in classification and localization, as discussed in this paper.

In TABLE \ref{table:Ablation}, the $mAP_{50}$ of the YOLOv10n baseline model is 37.3\%. The first method upsamples the feature maps using nearest neighbor upsampling before the minimum and medium resolution detection heads. The output is then fused with the feature maps before the maximum resolution detection head via the FMSA module. As a result, the performance of this single-detection-head network outperforms the three-detection-head YOLOv10n network. This suggests that for a detection head with the highest resolution, there is actually a relative lack of semantic information provided by deeper layers. After supplementing this missing information, the performance of a single detection head is comparable to that of three detection heads. \\
\indent The second method adds the FUS module in addition to FMSA module, performing global attention upsampling on the feature maps before the minimum and medium resolution detection heads of baseline network, instead of simple nearest neighbor upsampling, resulting in improved network performance. After incorporating global attention, recall rate (R) increased from 36.2\% to 37.1\%, indicating that the model has enhanced its ability to differentiate between objects and background. 
\begin{table}[t]
\centering
\caption{Comparison of different modules on VisDrone2019.}
\begin{tabular}{|c|ccc|c|c|c|c|}
\hline
\multirow{2}{*}{\textbf{No.}} & \multicolumn{3}{c|}{\textbf{Setting}} & \multicolumn{4}{c|}{\textbf{Metric}} \\
\cline{2-8}
 & \textbf{FMSA} & \textbf{FUS} & \textbf{FDS} & \textbf{P} & \textbf{R} & \textbf{mAP\textsubscript{50}} & \textbf{mAP\textsubscript{50-90}} \\
\hline
0 & \textcolor[gray]{0.80}{\ding{55}} & \textcolor[gray]{0.80}{\ding{55}} & \textcolor[gray]{0.80}{\ding{55}} & 48.6 & 36.7 & 37.3 & 22.4 \\
1 & \checkmark & \textcolor[gray]{0.80}{\ding{55}} & \textcolor[gray]{0.80}{\ding{55}} & 50.3 & 36.2 & 37.9 & 23.0 \\
2 & \checkmark & \checkmark & \textcolor[gray]{0.80}{\ding{55}} & 50.1 & 37.1 & 38.3 & 23.1 \\
3 & \checkmark & \textcolor[gray]{0.80}{\ding{55}} & \checkmark & 52.3 & 38.3 & 39.9 & 24.4 \\
4 & \checkmark & \checkmark & \checkmark & 52.9 & 40.3 & \textbf{41.7} & \textbf{25.0} \\
\hline
\end{tabular}
\label{table:Ablation}
\end{table}

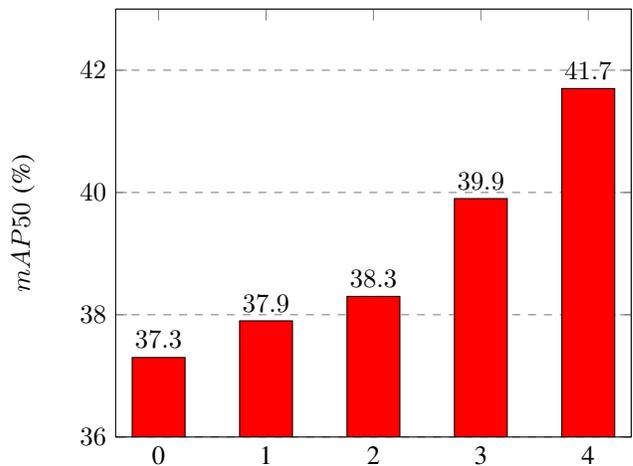
\begin{figure}[t]
    \centering
    \begin{tikzpicture}
        \begin{axis}[
            ybar,
            bar width=20pt,
            symbolic x coords={0, 1, 2, 3, 4},
            xtick=data,
            ylabel={$mAP{50}$ (\%)},
            ymin=36.0, ymax=43.0,
            nodes near coords,
            nodes near coords align={vertical},
            ymajorgrids=true, 
            grid style={dashed, gray}, 
            ytick align=inside,
            xtick align=inside,
            ]
            \addplot[fill=red] coordinates {
                (0, 37.3)
                (1, 37.9)
                (2, 38.3)
                (3, 39.9)
                (4, 41.7)
            };
        \end{axis}
    \end{tikzpicture}
    \caption{Comparison of different modules on VisDrone2019. 0 is the baseline model YOLO-v10; 1 is FMSA enabled; 2 is FMSA and FUS enabled; 3 is FMSA and FDS enabled; 4 is FMSA, FUS, and FDS enabled.}
    \label{fig:Ablation}
\end{figure}

The third method adds the FDS module on top of the first method, resulting in a significant performance boost. On the one hand, it demonstrates that introducing shallow layer features can significantly enhance the performance of the network. On the other hand, this also reveals the problem presented in this paper that the imbalance of classification and localization information exists. As the network architecture becomes deeper, the enhancement of semantic information is often accompanied by a decline in the representation of spatial information. Compared to the second method, in small target detection tasks, the benefit of spatial information in shallow layers slightly outweighs that of semantic information in deep layers.

The fourth method contains all the new modules FMSA, FUS, and FDS. It is the completed version of the proposed method. As shown in Fig.~\ref{fig:Ablation}, it is clear to see the performance enhancement with the proposed modules added. The experiment result demonstrates that the proposed approach effectively addresses the imbalance issue of classification and localization information in the network, which achieves optimal performance, with a $mAP_{50}$ of 41.7\%, a 4.4\% improvement over the baseline model.

\subsection{Visualization}  
\begin{figure}[t]
    \centering
    \includegraphics[width=1\linewidth]{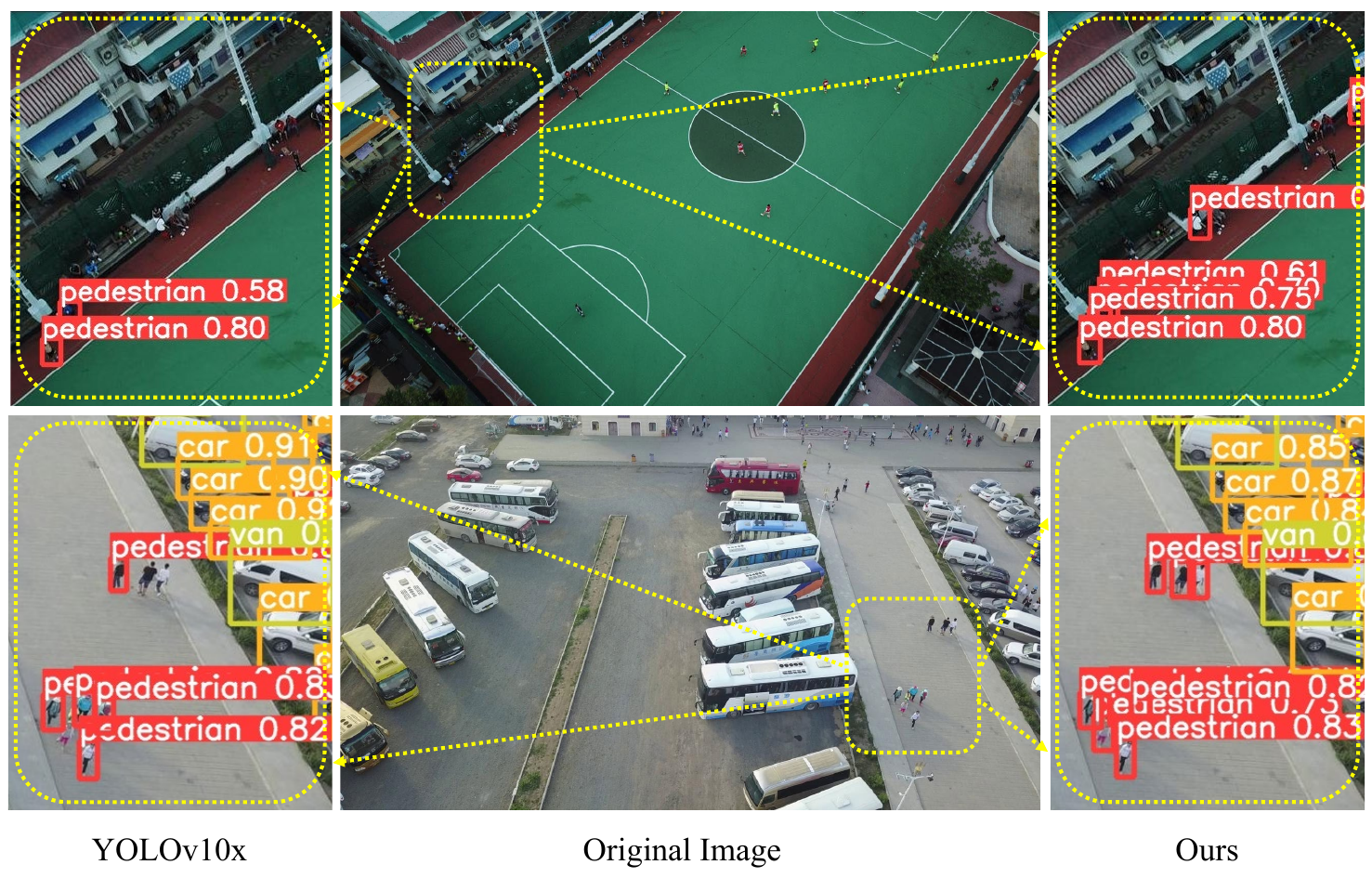}
\caption{Comparison of detection results on the VisDrone2019 dataset.}
\label{fig:Visualization}
\end{figure}
Finally, in Fig.\,\ref{fig:Visualization} we present representative visualizations from the VisDrone2019 dataset to provide an intuitive demonstration of the impact of our proposed method. These visualizations illustrate the model's enhanced detection capabilities, offering clear comparative evidence of its performance improvement over the baseline model.

The results demonstrate that our proposed approach achieves significant enhancement in detecting overlapping and extremely small targets across diverse scenarios compared to baseline model. This substantiates the method's improved precision and effectiveness in complex detection environments, particularly in addressing the challenges introduced by small and occluded objects. 

\section{Conclusion}\label{sec:con}
This paper addresses the imbalance between classification and localization information in CNN-based object detection networks, particularly in UAV scenarios. Existing methods, including multi-scale fusion and multi-detection-head designs like YOLO, partially alleviate this issue but often introduce redundancy or remain constrained by limited cross-layer connections.
We proposed a flexible feature fusion framework integrating FDS, FUS, and FMSA modules to enhance cross-layer connectivity and multi-scale fusion. This approach effectively improves the detection of small objects while maintaining computational efficiency. Experiments on the YOLO-v10 model demonstrate significant improvements in accuracy, especially for UAV-based detection tasks.
Despite these advancements, our method's performance on broader detection tasks and extremely high-resolution images warrants further exploration. Future work will focus on extending this framework to Transformer-based architectures and developing adaptive fusion techniques for diverse object detection scenarios.

\section*{Acknowledgement}
This work is supported by the Sci-Tech Innovation Initiative by the Science and Technology Commission of Shanghai Municipality (24ZR1419000),  and the National Science Foundation of China (12471501).

{\small 
\bibliographystyle{IEEEtran}
\bibliography{ref}
}

\end{document}